\documentclass[10pt,twocolumn,letterpaper]{article}

\usepackage[margin=1in]{geometry}
\usepackage[utf8]{inputenc}
\usepackage[T1]{fontenc}
\usepackage{times}
\usepackage{amsmath,amssymb}
\usepackage{booktabs}
\usepackage{graphicx}
\usepackage{adjustbox}
\usepackage{tikz}
\usetikzlibrary{arrows.meta,fit,backgrounds}
\usepackage{xcolor}
\usepackage{flushend}
\usepackage{placeins}  %
\usepackage[numbers,sort&compress]{natbib}
\usepackage[colorlinks=true,linkcolor=blue,citecolor=blue,urlcolor=blue]{hyperref}

\usepackage{llamapaper}
\usepackage[nameinlink,capitalize]{cleveref}

\graphicspath{{figures/}}

\newcommand{\ocria}{OCRia-nano-1}

\title{OCRia-nano-1: Fast, Accurate CPU Parsing with 8.1M Parameters}
\author{
  Logan Markewich \quad
  Simon Suo \\[4pt]
  \normalsize \texttt{\{logan, simon\}@runllama.ai}
}
\date{}

\LIPaperType{Technical report}
\LIPaperDate{November 2026}
\LIPDFAuthor{Logan Markewich, Simon Suo}
\LIPDFSubject{OCRia-nano-1: small local models for OCR, tables and layout in LiteParse}
\LIAbstract{Recent vision-language models achieve strong document parsing accuracy but are optimized for server-grade GPUs.
Existing CPU parsers still lag in accuracy, particularly in reconstructing tables and layout.
We present \ocria{}, an 8.1M-parameter document parsing pipeline optimized for CPU inference.
The pipeline combines native PDF extraction with small models for OCR and document structure.
LiteParse extracts available text and geometry, invokes OCR where needed, and supplies spatial features to LayoutNet,
a joint table and layout reconstruction model.
\ocria{} outperforms Indra and Marker on four of five benchmarks at comparable CPU throughput.
On ParseBench-2, it scores 59.2, compared with 55.1 for Indra and 52.5 for Marker.
It processes 27.8 pages per second on 16 vCPUs, twice the throughput of LiteParse's default Tesseract configuration.
Ablations show that joint table and layout reconstruction accounts for much of the accuracy gain,
while larger OCR models provide only modest additional improvements.
We release the model weights on Hugging Face under a modified OpenRAIL license, with LiteParse integration as an opt-in OCR backend.}
\LIPaperLinks{%
  \href{https://github.com/run-llama/liteparse}{Code}
  \hspace{1.1em}
  \href{https://huggingface.co/llamaindex/OCRia-nano-1}{Models}
}
\LIHero{%
  \begin{center}
  \includegraphics[width=\textwidth]{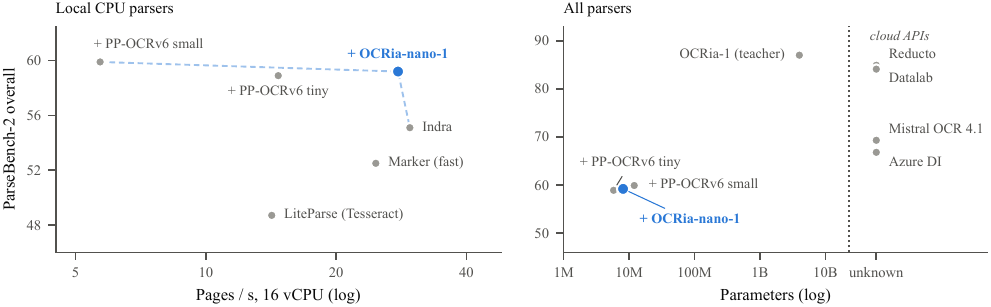}
  \captionof{figure}{Parsing accuracy versus CPU throughput (left) and model size (right).
  Accuracy is the overall ParseBench-2 score. Throughput is measured in pages per second on 16 vCPUs.
  The dashed line marks the best speed--accuracy tradeoffs among the compared CPU parsers.
  The PP-OCRv6 points run LiteParse and LayoutNet with PP-OCRv6 in place of our OCR models (\cref{tab:efficiency,tab:ablations}).
  Cloud APIs with undisclosed model sizes are shown separately.}
  \label{fig:accuracy_cost}
  \end{center}
}

\begin{document}

\maketitle

\begin{figure*}[t]
\centering
\begin{tikzpicture}[
  x=0.93cm, y=1cm,
  font=\scriptsize,
  >={Stealth[length=1.6mm]},
  stage/.style={draw=black!45, rounded corners=1.5pt, align=center, inner sep=2.5pt,
                minimum height=8mm, text width=17mm, fill=black!4},
  model/.style={stage, draw=lpblue, fill=lpblue!10},
  io/.style={align=center, text width=13mm},
  flow/.style={->, draw=black!60, thick},
]
\definecolor{lpblue}{HTML}{2A78D6}

\node[io]    (pdf)    at (0, 0)    {PDF / image};
\node[stage] (pdfium) at (2.2, 0)  {PDFium\\text layer};
\node[stage] (merge)  at (9.0, 0)  {OCR / text-layer\\merge};
\node[stage] (proj)   at (11.3, 0) {Grid projection};
\node[model] (tn)     at (13.6, 0) {LayoutNet {\tiny(4.3M)}\\tables, block classes, blocks};
\node[io]    (out)    at (15.8, 0) {Markdown / JSON / blocks};

\node[stage] (sel)    at (2.2, -1.6) {Page check:\\OCR needed?};
\node[model] (det)    at (4.5, -1.6) {Line detector\\{\tiny 0.08M}};
\node[model] (router) at (6.7, -1.6) {Script router\\{\tiny 0.04M}};
\node[model, text width=19mm] (rec) at (9.0, -1.6) {CTC recognizers\\Latin / World / RTL / CJK\\{\tiny 0.71--1.27M each}};

\draw[flow] (pdf) -- (pdfium);
\draw[flow] (pdfium) -- node[above, font=\tiny, text=black!60] {native text} (merge);
\draw[flow] (pdfium) -- (sel);
\draw[flow] (sel) -- (det);
\draw[flow] (det) -- (router);
\draw[flow] (router) -- (rec);
\draw[flow] (rec) -- (merge);
\draw[flow] (merge) -- (proj);
\draw[flow] (proj) -- (tn);
\draw[flow] (tn) -- (out);

\begin{scope}[on background layer]
  \node[draw=lpblue!60, dashed, rounded corners=3pt, inner sep=1.6mm, fit=(det)(router)(rec),
        label={[font=\scriptsize\itshape, text=lpblue]below:ocria-nano OCR}] {};
\end{scope}
\end{tikzpicture}
\caption{The \ocria{} document parsing pipeline. LiteParse extracts native PDF text and invokes OCR where needed.
It merges native and recognized text and derives spatial features for LayoutNet to reconstruct tables and layout.
Blue boxes show learned models, and grey boxes show LiteParse processing stages.}
\label{fig:hero}
\end{figure*}
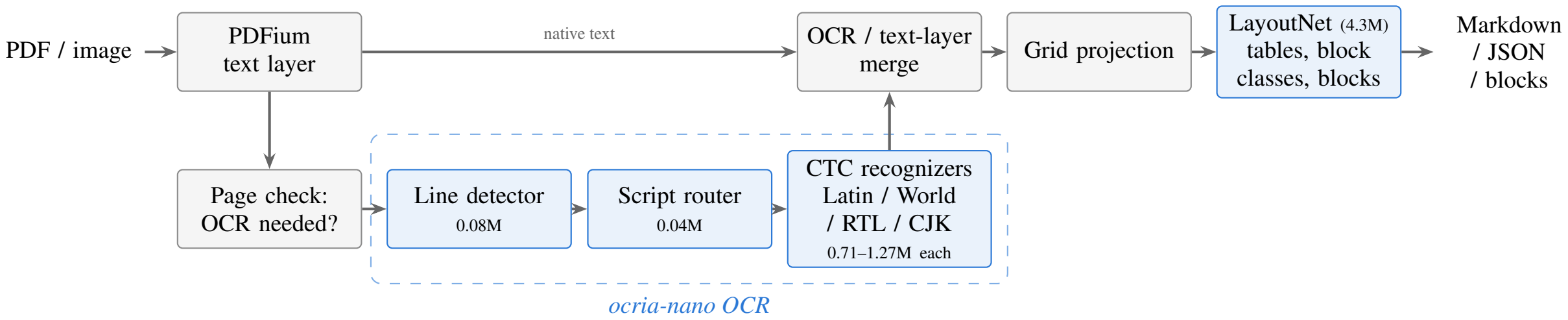

\section{Introduction}
\label{sec:intro}

Converting a document to a structured representation like Markdown typically covers a broad spectrum of complexity,
from simple paragraphs to complex forms and nested tables. This spectrum leads to a diverse landscape of approaches
to document parsing, from single-shot VLM systems that process purely from screenshots and leverage heavy GPU resources \cite{nougat,gotocr2,olmocr,mineru25},
to heuristic-based systems tuned for running as fast as possible on CPU \cite{pymupdf4llm,pdfplumber}. This leads to a
common decision point for users and consumers of these systems: whether to run accurate models at the cost of compute,
or decide that the data is straightforward or latency-sensitive enough for CPU-based systems.

Document parsers have largely taken the compute path: recent document parsers and OCR systems rely on vision-language models (VLMs), typically in the 1--4B parameter range.
OCRia-1 (4B parameters) \citep{ocria1}, LightOnOCR-3 (0.8B, 1B, and 4B parameter versions) \citep{lightonocr3,lightonocr} and NVIDIA Nemotron 3.5 Parse (0.6B parameters) \citep{nemotronparse} are all recent
models that achieve strong parsing accuracy and are optimized for server-grade GPUs and accelerators.
Benchmarks like ParseBench-2 \citep{parsebench2} and olmOCR-bench \citep{olmocr,olmocr2} help categorize the strengths and weaknesses of these models.
At the same time, smaller CPU-based systems have seen less development. The most notable is Paddle's PP-OCRv6 (1.5M to 34.6M parameters) \citep{ppocrv6}, although as a classical OCR system, it
does not produce markdown without additional heuristics. Indra \citep{indra} is a 12M parameter pipeline with a similar architecture, whose additional models for page orientation, reading order
and math, combined with heuristics, enable markdown output.

In this paper, we introduce \ocria{}, an 8.1M-parameter document parsing pipeline optimized for CPU inference.
\ocria{} combines native PDF extraction with small models for OCR and document structure. LiteParse \citep{liteparse}
extracts available text and its positions, invokes OCR where text is missing or unusable, and derives spatial cues from the page.
LayoutNet uses these inputs to reconstruct tables and layout, allowing the pipeline to organize the document into Markdown
with compact models.
The models integrate with LiteParse as an opt-in replacement for its default Tesseract backend, improving both throughput and accuracy.

\ocria{} outperforms Indra and Marker on four of five benchmarks at comparable CPU throughput.
On ParseBench-2, it scores 59.2, compared with 55.1 for Indra and 52.5 for Marker.
It processes 27.8 pages per second on 16 vCPUs, twice the throughput of LiteParse's default Tesseract configuration
(\cref{fig:accuracy_cost}). Ablations show that joint table and layout reconstruction accounts for much of the accuracy gain,
while larger OCR models provide only modest additional improvements.

Our contributions are:
\nopagebreak[4]

\begin{enumerate}
  \interlinepenalty=10000
  \item \textbf{An accurate, efficient CPU document parser}: \ocria{} combines LiteParse's native text and geometry extraction, selective OCR, and joint table and layout reconstruction in an 8.1M-parameter CPU pipeline.
  \item \textbf{A component-level analysis of CPU document parsing}: through ablations and model-scaling experiments, we quantify the accuracy and efficiency effects of individual components. We show that learned table and layout reconstruction delivers larger parsing gains than scaling OCR models.
  \item \textbf{Released models for practical use}: we release the OCR and LayoutNet weights on Hugging Face, with LiteParse integration for local CPU document parsing.
\end{enumerate}
\section[A modular document parsing pipeline optimized for CPUs]{\raggedright A modular document parsing pipeline \mbox{optimized for CPUs}}
\label{sec:system}

\ocria{} is a document parsing pipeline optimized for CPU inference. Its design combines rich text and geometry
extraction with small models for OCR and document structure. Native PDF text provides both content and spatial
information, while OCR supplies text where the native layer is missing or unusable. LiteParse, the pipeline's PDF
processing engine, uses a grid projection to derive further cues about how text is aligned and grouped on the page.
We use this information as the foundation
for a lightweight model that reconstructs tables and layout.

The pipeline has three components (\cref{fig:hero}). LiteParse extracts and merges
native and OCR text, computes spatial features, and assembles the final Markdown. The OCR models detect text lines
and recognize their contents. LayoutNet uses the extracted text, geometry, and LiteParse's spatial features to
predict layout classes and relationships between text units, with a small image branch providing visual context.
Together, these components recover both the document's text and the structure needed to organize it into Markdown.
\Cref{tab:models} lists the learned components.

\begin{table}[t]
\centering\small
\adjustbox{max width=\linewidth}{%
\begin{tabular}{@{}llr@{}}
\toprule
Model & Role & Params \\
\midrule
\texttt{det} & Text-line detector & 0.08M \\
\texttt{router} & Script router & 0.04M \\
\texttt{latin\_rec} & Latin recognizer & 0.80M \\
\texttt{world\_rec} & World-script recognizer & 0.98M \\
\texttt{rtl\_rec} & RTL recognizer & 0.71M \\
\texttt{cjk\_rec} & CJK recognizer & 1.27M \\
\texttt{layout\_net} & Joint table + layout model & 4.27M \\
\midrule
Total & & 8.1M \\
\bottomrule
\end{tabular}%
}
\caption{The \ocria{} pipeline of models. Parameter counts from the released model weights.}
\label{tab:models}
\end{table}

\subsection{LiteParse: extraction and orchestration}
\label{sec:liteparse}

LiteParse is a PDF processing engine that mines text and geometry information from documents
and uses it to create layout projections and markdown \cite{liteparse}. It orchestrates the OCR and LayoutNet models
and merges OCR text with native PDF text while ensuring OCR is only applied on pages where it is needed.

LiteParse runs page checks to decide
which pages need OCR, triggering for scans, image-only regions, and invisible or garbled text layers.
On mixed pages, text the PDF already provides is masked out of the page render,
so the OCR models read only what the text layer is missing.

LiteParse's grid projection places the text from the PDF and OCR on an ASCII grid that mimics the page's original spacing and layout.
As a side-effect of this process, a wealth of statistics are gathered about the document. We select eight signals from the grid projection
as inputs to LayoutNet, alongside the text geometry and page image (\cref{sec:layoutnet}).
LiteParse supplies text and geometric cues, while LayoutNet learns table and layout decisions that are difficult to capture with fixed rules.

\subsection{OCR: text detection and recognition}
\label{sec:ocr}

The OCR models form a pipeline where a detector finds text lines, each line is cropped
and straightened, and a CTC recognizer \citep{graves2006ctc,shi2017crnn} reads the crop. PP-OCR
\citep{ppocr,paddleocr} popularised this pipeline, and routing lines to script-specific recognizers is
common in multilingual OCR (\cref{sec:related}). Our six OCR models in
\cref{tab:models} total 3.86M parameters.

\Cref{sec:eval:ablations} compares \ocria{} with other OCR models swapped into the same pipeline.

\paragraph{Line detector.}

The detector is a small segmentation network from the DBNet family \citep{liao2020dbnet}. It predicts a
single map of shrunken line cores, and connected components of that map grow back into rotated
quads. The shipped detector has 82k parameters and is built almost entirely from depthwise-separable convolutions.

\paragraph{Script router and recognizers.}

A 39k-parameter router reads a text line crop and jointly predicts the script family (Latin, CJK,
right-to-left, or ``world'') and whether the line is upside down. The
page takes a majority script, and individual lines may override it when the router is confident (\cref{fig:ocr_routing}). The
upside-down votes rotate the whole page before recognition.

Each script family has its own recognizer, and all four are the same convolution-only CTC architecture.
Greedy decoding happens inside the exported graph, so the
runtime only maps indices to characters. The models range from 0.71M parameters (RTL) to 1.27M (CJK, which
carries the largest character set). A page of another script sent to the Latin recognizer reads as confident-looking nonsense, so when a page's
Latin reading is unusually unsure, a sample of its lines is re-read by every recognizer, and the
page moves to another recognizer if that one is clearly more confident. A line-level version of the same check catches
mixed pages, such as English labels on an Amharic map.
\begin{table*}[t]
\centering\small
\adjustbox{max width=\textwidth}{%
\begin{tabular}{@{}lllrl@{}}
\toprule
Type & Source & Label & Size & Used by \\
\midrule
\multicolumn{5}{@{}l}{\emph{OCR}} \\
PDF text layer      & web-crawled PDFs                        & PDFium text, PP-OCRv6 must agree        & 35.2k pages  & det, all rec., router \\
Scans               & scanned pages from the same crawl       & PP-OCRv6 text, a second reader must agree & 10.1k pages & det, Latin \\
Synthetic text      & Wikipedia sentences, 47 languages       & exact                                    & 30.0k pages  & det, all rec., router \\
Grounded corpus     & OCRia-1 SFT corpus, 6 sets              & text + word boxes                        & 11.1k pages  & det, Latin, RTL, world \\
OCR layers          & grounded scans with an invisible OCR layer & layer text, corpus text must agree    & 4.8k crops   & world \\
Handwriting         & Wikimedia Commons                       & PP-OCRv6 box + OCRia-1 text              & 2.2k pages   & Latin \\
Typewriting         & CIA CREST (Internet Archive mirror)     & Tesseract box + OCRia-1 text             & 400 pages    & Latin \\
Table cells, IDs    & SFT-corpus tables; synthetic            & PDF text; exact                          & 658k crops   & Latin \\
Public              & HierText, GNHK, CORD, CORU, CATMuS      & human                                    & 381k crops   & det, Latin, RTL \\
\midrule
\multicolumn{5}{@{}l}{\emph{LayoutNet (tables + layout)}} \\
Corpus tables       & OCRia-1 SFT corpus, 9 sets              & HTML tables grounded to units            & 18.3k pages  & LayoutNet \\
PMC                 & PMC Open Access (CC BY, BY-SA, CC0)     & OCRia-1 tables (direct call), text match & 7.0k pages   & LayoutNet \\
Layout              & OCRia-1 SFT corpus, 18 sets             & layout blocks, 13 classes                & 23.7k pages  & LayoutNet \\
\bottomrule
\end{tabular}}
\caption{Training data sources. Sizes count usable pages or crops after filtering.
The detector trains on the text-layer, scan, synthetic, grounded, HierText and GNHK lanes. The
corpus-table and PMC pages are each built twice, once with PDF-text units and once with OCR units.}
\label{tab:data_lanes}
\end{table*}

\subsection[LayoutNet: joint table and layout reconstruction]{\raggedright LayoutNet: joint table and layout \mbox{reconstruction}}
\label{sec:layoutnet}

LayoutNet is a joint table and layout detection model.
LayoutNet produces several predictions in a single pass for each text unit, namely is it in a table (and which row, column and cell),
which block does it belong to, and what class is that block. Text-unit
transformers using pairwise heads are not new, and graph and pairwise-relation methods have a long history in
table structure recognition (\cref{sec:related}). What makes LayoutNet unique is the tight integration
with LiteParse, using outputs and measurements from LiteParse's grid projection algorithm, along with
typical text information, geometry, and page images.

\paragraph{Units and inputs.}

The inputs to LayoutNet are page screenshots and \emph{units}. Units are runs of text split
heuristically where a table cell or layout block could have a boundary. Unit texts themselves
come from either embedded PDF data or OCR results.

Each unit also supplies its geometry and position, its text shape (digits,
case, punctuation, length), and its style when available from the PDF. During training,
we mask styles so that units from OCR stay in-distribution. LiteParse itself adds an additional
eight features from the grid projection algorithm (\cref{sec:liteparse}). They are a flag for whether the projection placed the unit,
whether the text flows as prose, how many text items share its left, right and centre alignment, and which of those three anchors it
snapped to, with the text unit's projection block and XY-cut leaf passed alongside.
Detected ruling lines also provide further geometric information to the text unit.

A small CNN image branch gives the model visual features. A strided CNN turns a greyscale page
screenshot into patch tokens, and each unit samples the feature map under its box. The top encoder
layers cross-attend to the image tokens, carrying visual features through the model.

\paragraph{Model and decoding.}

The encoder is a 6-layer pre-norm transformer at width 256. \emph{Node} heads predict
whether a unit is in a table or header. \emph{Pair} heads predict pairwise relations,
whether units share a table, row, column, cell, or layout block. A \emph{class} head assigns
one of our 13 layout classes (text, heading, title, page header and footer, list, table, caption, footnote, formula, code,
figure text and key-value). In total with the image CNN branch, LayoutNet amounts to 4.3M parameters.

We decode tables by clustering the same-table prediction graph.
Rows, columns, and cells come from the clusters inside each table. Some post-processing works
to remove false positives like tables of contents, chart axes, and other table-like structures. These elements are
not tables, and rendering them as tables makes the markdown harder to read. Layout blocks are connected components of the same-block graph, with each taking the class
with the largest summed probability (\cref{fig:layout_blocks}).

LayoutNet's tables claim their lines first. LiteParse's heuristic table detectors then run on the remaining lines, which
recovers tables the model misses, but they also read aligned prose, stat callouts and timelines as grids. We therefore keep a
heuristic table only when LayoutNet's class head agrees that its lines are tabular (mean table probability at least 0.3).
Otherwise, its lines fall back to paragraphs. We evaluate this filter in \cref{sec:eval:ablations}.
\section{Training data and supervision}
\label{sec:data}

\ocria{} is a pipeline of models, and requires unique training data for each component. The detector
learns from line boxes, transcribed line crops train the recognizers, and page-level
layout annotations train our LayoutNet. Additional data and training details are in \cref{app:data,app:training}.

\begin{table*}[t]
\centering\small
\adjustbox{max width=\linewidth}{%
\begin{tabular}{@{}lrrrrrr@{}}
\toprule
System & ParseBench-2 & OmniParseBench & ParseBench-1 & olmOCR-bench & BHL CER $\downarrow$ & pages/s \\
\midrule
\multicolumn{7}{@{}l}{\emph{Local, CPU (run by us)}} \\
LiteParse (Tesseract) & 48.7 & 42.3 & 40.5 & 45.7 & 0.275 & 14.2 \\
Indra 1.1.2 & 55.1 & 70.9 & 42.6 & 60.8 & 0.062 & \textbf{29.6} \\
Marker (fast) & 52.5 & 69.8 & 40.4 & 60.8 & 0.067 & 24.7 \\
\textbf{LiteParse + \ocria{}} & \textbf{59.2} & \textbf{74.8} & \textbf{52.7} & \textbf{60.9} & \textbf{0.048} & 27.8 \\
\midrule
\multicolumn{7}{@{}l}{\emph{GPU / cloud (published numbers)}} \\
OCRia-1 (teacher) & 87.0 & -- & 82.8 & 84.6 & -- & -- \\
Chandra 2 & 78.9 & 90.1 & 70.6 & 85.8 & 0.029 & -- \\
Infinity-Parser2-Pro & 78.5 & 87.0 & 71.2 & 87.6 & -- & -- \\
LightOnOCR-3 4B & 78.4 & 82.9 & 75.3 & 86.3 & 0.031 & -- \\
Mistral OCR 4.1 & 69.3 & 89.5 & -- & 81.9 & -- & -- \\
\bottomrule
\end{tabular}%
}
\caption{Accuracy and CPU throughput (16 vCPUs). Best CPU results in bold, including ties.}
\label{tab:main_results}
\end{table*}

\paragraph{Data sources and coverage.}
We reuse annotations from permissive public datasets and OCRia-1's supervised fine-tuning corpus.
Public training data for document OCR and table structure
is scarce and often licensed for non-commercial use only \citep{marti2002iam,jaume2019funsd}, so we generate additional
labels automatically.

We also obtain labels from PDF text layers, synthetic pages with known text, and direct predictions from
OCRia-1 \citep{ocria1}, our 4B teacher model. We reuse table and layout annotations from LlamaParse's agentic pipeline
and ground them using layout blocks and word boxes.

PDF data and OCRia-1 give high quality transcriptions, with OCR agreement helping to supply text geometry
when it is not available. \Cref{tab:data_lanes} summarizes the training data and supervision for each model.

\paragraph{OCR supervision.}
We rely on an agreement filter to verify training data: an independent reader (PP-OCRv6 \citep{ppocr,paddleocr}, or Tesseract)
reads each line again, and a label is kept only when the two agree. For image-born documents, labels include existing
annotations and teacher transcriptions,
while PDF text segments are used when available. Text lines are kept
when the readers agree to within a character error rate of 0.15. We set this threshold by inspecting crops sorted by disagreement, where
we found this to be an acceptable balance between readers. On the PDF text-layer data, this filter drops 3\% of the transcribed Latin text, while dropping
63\% of the right-to-left and 53\% of the world script lines, due to weak off-the-shelf OCR on these texts.
Synthetic texts make up the bulk of our non-Latin text to compensate for the impact of filtered data.

Most labelled pages are born-digital, so training pages pass through an online scan simulator that teaches the detector
scan conditions from them, applying perspective warps, paper texture, uneven lighting,
fax binarisation, show-through, blur and compression at random. Non-text distractors (photos, halftones,
barcodes) are also pasted into empty regions as negatives, which cuts false detections on pictures and barcodes.

\paragraph{Table and layout supervision.}

Training LayoutNet requires assigning table structure and layout labels to the text units extracted by LiteParse.
Table annotations from the reused corpus and the teacher are represented as HTML, which
provides row and column locations but not box locations.
The text units, which come from either the PDF data or our OCR, have the boxes but no
table structure. To connect these two, we assign each text unit to an annotated cell using
the overlap of word boxes and matching text. Once text units are assigned to annotated cells, these assignments define
which pairs share a table, row, column, or cell. Layout classes are labelled the same way, where
overlap determines the class of a text unit, assigning one of our 13 layout classes (\cref{sec:layoutnet}).
We also drop chart values OCRia-1 transcribes as tables, since they can't
be grounded to the input.
\section{Evaluation}
\label{sec:eval}

\begin{table*}[t]
\centering\small
\adjustbox{max width=\linewidth}{%
\begin{tabular}{@{}lrrrrrrr@{}}
\toprule
System & Overall & Text & Formatting & Tables & Charts & Layout & Forms \\
\midrule
\multicolumn{8}{@{}l}{\emph{Local, CPU (run by us)}} \\
LiteParse (Tesseract) & 48.7 & 58.8 & 78.0 & 36.2 & 1.2 & 49.4 & 68.5 \\
Indra 1.1.2 & 55.1 & 68.4 & 76.1 & 47.6 & \textbf{3.1} & 65.2 & \textbf{70.2} \\
Marker (fast) & 52.5 & 62.5 & 75.5 & 43.6 & \textbf{3.1} & 61.0 & 69.2 \\
\textbf{LiteParse + \ocria{}} & \textbf{59.2} & \textbf{68.5} & \textbf{82.2} & \textbf{59.5} & \textbf{3.1} & \textbf{72.4} & 69.8 \\
\midrule
\multicolumn{8}{@{}l}{\emph{GPU / cloud (published numbers)}} \\
OCRia-1 (teacher) & 87.0 & 87.0 & 92.7 & 89.9 & 77.4 & 89.2 & 85.6 \\
Reducto & 84.9 & 81.8 & 82.6 & 84.9 & 89.1 & 86.0 & 85.2 \\
Datalab (accurate) & 84.1 & 91.1 & 92.5 & 90.4 & 67.8 & 82.4 & 80.6 \\
Chandra 2 & 78.9 & 77.2 & 88.8 & 85.6 & 62.5 & 80.6 & 78.7 \\
MinerU2.5-Pro & 76.9 & 77.1 & 79.0 & 80.6 & 64.0 & 86.2 & 74.5 \\
PaddleOCR-VL 1.6 & 71.5 & 73.3 & 79.4 & 73.0 & 57.5 & 74.5 & 71.2 \\
Mistral OCR 4.1 & 69.3 & 85.0 & 81.9 & 81.8 & 1.2 & 85.9 & 79.7 \\
Azure DI Layout & 66.8 & 79.8 & 77.8 & 77.6 & 1.5 & 83.9 & 80.3 \\
\bottomrule
\end{tabular}%
}
\caption{Overall and per-dimension ParseBench-2 scores. Best CPU results in bold, including ties.}
\label{tab:pb2}
\end{table*}

This section answers three questions: how accurate \ocria{} is against various local and hosted parsers (\cref{sec:eval:main}),
how fast it runs on CPU (\cref{sec:eval:efficiency}), and which of its parts contribute accuracy (\cref{sec:eval:ablations}).

\subsection{Benchmarks and protocol}
\label{sec:eval:protocol}

Our primary benchmark is ParseBench-2 \citep{parsebench2}, measuring accuracy on text, formatting, tables, charts,
layout grounding, and forms across real business documents. We also report
OmniParseBench \citep{omniparsebench}, ParseBench-1 \citep{parsebench1}, olmOCR-bench \citep{olmocr,olmocr2}, and the historical
print BHL IMPACT dataset \citep{bhl_impact,bhl_leaderboard} for OCR character error rates (CER).

We compare \ocria{} to the CPU document parsers Indra \citep{indra} and Marker \citep{marker}, and to LiteParse's
default configuration with Tesseract 5 \citep{smith2007tesseract}. We also compare PP-OCRv6 \citep{ppocrv6} as a
replacement for our OCR models within the same pipeline (\cref{tab:ablations}).
\Cref{tab:main_results,tab:pb2} include GPU and cloud systems as reference points.

The CPU results in \cref{tab:main_results,tab:pb2} are from our runs, using LiteParse's release build where
applicable. GPU and cloud results come from published outputs and benchmark reports. All ParseBench-2 outputs are
scored with the same official scorer. Benchmark versions, scoring details, baseline configurations, and per-benchmark
settings are in \cref{app:eval}.

\subsection{Main results}
\label{sec:eval:main}

\Cref{tab:pb2} shows that \ocria{} closes the gap with commercial parsers and beats the other local CPU parsers on ParseBench-2.
It adds 10.6 points to the default LiteParse configuration with Tesseract for OCR. For systems that
can run locally on CPU, \ocria{} scores the biggest lead in tables. Indra is ahead of \ocria{} only on forms (70.2 against
69.8), and the two are level on text (68.5 against 68.4).
\ocria{}'s lead comes from tables (59.5 against 47.6), layout grounding (72.4 against 65.2) and formatting (82.2 against 76.1). \ocria{} does
not ship with chart understanding and trails every other VLM and specialized parser that supports chart-to-table
conversion.

On OmniParseBench's official harness (\cref{tab:main_results}), LiteParse with \ocria{} scores 74.8
(text 59.3, tables 74.3, layout 90.9), against 42.3 for the default configuration.
The tables family accounts for most of the gain since the default configuration has no table model and scores 14.5 there,
while LayoutNet brings it to 74.3.

\begin{figure}[t]
\centering
\includegraphics[width=\columnwidth]{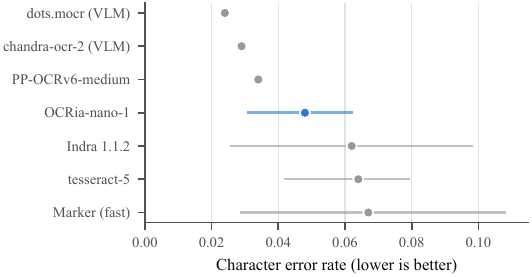}
\caption{Character error rate on 2,165 historical book pages from BHL IMPACT. Lower is better.
Error bars show 95\% confidence intervals where available.}
\label{fig:bhl}
\end{figure}

On BHL's historical print (\cref{fig:bhl}), \ocria{} reaches a character error rate of 0.048,
while Tesseract 5 scores 0.064 on the leaderboard; LiteParse's default Tesseract configuration scores 0.275 in our run. The VLMs on the leaderboard reach 0.024--0.031, and
PP-OCRv6-medium (34.6M parameters, 9$\times$ our 3.9M OCR stack) reaches 0.034. Our run of Indra 1.1.2 and Marker 2.2.0 score 0.062 and 0.067 respectively.
On ParseBench-1 and olmOCR-bench (\cref{tab:main_results}), \ocria{} adds 12.2 and 15.2 points to
the default Tesseract configuration. olmOCR-bench moves more than ParseBench-1 because it expects
page furniture (running headers, footers and page numbers) to be removed. Its headers-and-footers category rises from 50.8\%
with the default configuration to 88.2\% with \ocria{}.

\subsection{Inference efficiency}
\label{sec:eval:efficiency}

\ocria{} has 8.1M parameters, roughly 500$\times$ fewer than OCRia-1, and it runs on CPU only.
The OCR models run through ONNX Runtime, in-process. LayoutNet runs through ONNX Runtime as well, except on macOS,
where a plain-Rust implementation of the same weights is faster.
The installed package is 59.9\,MB on Linux and 75.3\,MB on macOS, of which 30\,MB are the models and the rest is ONNX Runtime.

\paragraph{Benchmark setup.}
\Cref{tab:efficiency} measures throughput on a 16-vCPU cloud container across five hosts, for
LiteParse's default Tesseract settings (with and without LayoutNet), Indra, Marker, LiteParse with PP-OCRv6 tiny and
small in place of our OCR models, and LiteParse with \ocria{} (with and without LayoutNet).
The PP-OCR configurations run the same pipeline as
\cref{tab:ablations}, including LayoutNet, with one RapidOCR server per worker on the worker's core.
PP-OCR runs through LiteParse's HTTP OCR interface, which adds a small per-page request cost.
When enabled, LayoutNet runs on every page since it produces the layout blocks as well as the tables.
The workload is 300 random pages from ParseBench-1, 60 from each of its five categories (tables,
charts, layout, text content and text formatting). Almost all are PDFs, a mix of born-digital pages and
scans.

\paragraph{Throughput and latency.}
\ocria{} parses 27.8 pages per second. That is 2.0$\times$ the default configuration, 1.9$\times$
PP-OCRv6 tiny and 4.8$\times$ PP-OCRv6 small, and it is within 0.7 ParseBench-2 points of the PP-OCRv6 small
model (\cref{tab:efficiency}). On one core, an average page takes 426\,ms with
\ocria{}, 751\,ms with Tesseract, and 773\,ms with PP-OCRv6 tiny. At P90,
pages finish within 0.9\,s, compared to 2.4\,s with Tesseract and 1.6\,s with PP-OCRv6 tiny.
Indra runs at 29.6 pages per second and an average of 440\,ms, landing within machine variation when compared with \ocria{} while
performing worse on accuracy benchmarks except for a near-tie on olmOCR-bench (\Cref{tab:main_results}).
Marker's bundled Indra predictor defaults to 4 ONNX threads, which oversubscribes during the throughput benchmark. Adjusting
this to one thread per worker to match the other configurations parses 24.7 pages per second (0.9$\times$ \ocria{}).

\paragraph{Component costs and optimizations.}
\Cref{tab:efficiency} separates costs by running the same OCR with and without LayoutNet.
LayoutNet adds about 100\,ms per page on one core (426 against 325\,ms mean latency). Without it, \ocria{}'s OCR parses 37.2 pages per
second, more than two and a half times the Tesseract default.

These latency and throughput numbers were achieved using several optimizations across the pipeline and models.
On mixed pages we mask the text the PDF already provides before OCR, PNG
image inputs enter the PDF without re-encoding, and \ocria{} renders pages at a default 250 dpi,
which in our tests matched 300 dpi on benchmarks with about a third fewer pixels to process.

\begin{table}[t]
\centering\small
\adjustbox{max width=\linewidth}{%
\begin{tabular}{@{}lrrrr@{}}
\toprule
 & PB-2 & Pages/s & \multicolumn{2}{c}{ms per page, one core} \\
\cmidrule(l){4-5}
Configuration & overall & 16 vCPU & mean & p90 \\
\midrule
LiteParse (Tesseract) & 48.7 & 14.2 $\pm$ 1.6 & 751 & 2,420 \\
\quad + LayoutNet & 56.9 & 12.8 $\pm$ 1.4 & 859 & 2,626 \\
Indra 1.1.2 & 55.1 & 29.6 $\pm$ 5.3 & 440 & 871 \\
Marker (fast) & 52.5 & 24.7 $\pm$ 3.5 & 578 & 1,111 \\
LiteParse + PP-OCRv6 tiny + LayoutNet & 58.9 & 14.7 $\pm$ 2.7 & 773 & 1,568 \\
LiteParse + PP-OCRv6 small + LayoutNet & 59.9 & 5.7 $\pm$ 0.4 & 1,963 & 4,988 \\
LiteParse + \ocria{} OCR only & 51.2 & 37.2 $\pm$ 4.8 & 325 & 751 \\
\textbf{LiteParse + \ocria{} (release)} & 59.2 & 27.8 $\pm$ 4.4 & 426 & 862 \\
\bottomrule
\end{tabular}%
}
\caption{ParseBench-2 accuracy and CPU inference speed by configuration.}
\label{tab:efficiency}
\end{table}

\subsection{Ablations}
\label{sec:eval:ablations}

\begin{table*}[t]
\centering\small
\adjustbox{max width=\linewidth}{%
\begin{tabular}{@{}lrrlrrr@{}}
\toprule
\emph{OCR models (LayoutNet on)} & Params & Overall & $\Delta$ overall [95\% CI] & Text &  &  \\
\midrule
none (text layer only) & -- & 53.7 & $-5.5$ {\scriptsize[$-6.2$, $-4.9$]} & 57.1 & & \\
Tesseract 5, 300 dpi & -- & 56.9 & $-2.4$ {\scriptsize[$-2.8$, $-1.9$]} & 59.2 & & \\
PP-OCRv5 mobile & 3.1M & 58.7 & $-0.5$ {\scriptsize[$-0.8$, $-0.2$]} & 66.7 & & \\
PP-OCRv6 tiny & 1.5M & 58.9 & $-0.4$ {\scriptsize[$-0.7$, $-0.1$]} & 66.6 & & \\
\textbf{\ocria{}} & 3.9M & \textbf{59.2} & (release) & 68.5 & & \\
PP-OCRv6 small & 7.7M & 59.9 & $+0.7$ {\scriptsize[$+0.4$, $+1.0$]} & 71.9 & & \\
PP-OCRv6 medium & 34.6M & 60.2 & $+0.9$ {\scriptsize[$+0.6$, $+1.2$]} & 72.3 & & \\
\ocria{}, no LayoutNet & 3.9M & 51.2 & $-8.1$ {\scriptsize[$-8.6$, $-7.5$]} & 68.1 & & \\
\midrule
\emph{LayoutNet variants} & Params & Overall & $\Delta$ overall [95\% CI] & Tables & Layout &  \\
\midrule
Tables only (no layout heads, d192) & 1.9M & 55.3 & $-3.9$ {\scriptsize[$-4.3$, $-3.4$]} & 58.1 & 53.0 & \\
Joint (tables + layout, d192) & 1.9M & 58.3 & $-1.0$ {\scriptsize[$-1.3$, $-0.7$]} & 56.4 & 69.7 & \\
Joint, wider (d256, layout loss $\times$0.5) & 3.4M & 58.2 & $-1.1$ {\scriptsize[$-1.4$, $-0.7$]} & 57.7 & 69.4 & \\
\textbf{+ image branch (shipped)} & 4.3M & \textbf{59.2} & (release) & 59.5 & 72.4 & \\
+ capacity (d384, 8 layers) & 11.6M & 59.2 & $\pm0.0$ {\scriptsize[$-0.3$, $+0.2$]} & 59.2 & 71.9 & \\
\midrule
\emph{Latin recognizer} & Params & Overall & $\Delta$ overall [95\% CI] & Text & HierText CER & ms/line \\
\midrule
slim, from scratch & 0.76M & 58.9 & $-0.4$ {\scriptsize[$-0.6$, $-0.2$]} & 67.0 & 0.162 & 2.9 \\
large, from scratch & 4.26M & 59.3 & $+0.1$ {\scriptsize[$-0.1$, $+0.2$]} & 68.4 & 0.127 & 10.5 \\
xl, from scratch & 9.56M & 59.5 & $+0.3$ {\scriptsize[$+0.1$, $+0.4$]} & 69.4 & 0.116 & 21.6 \\
shipped, fp32 & 0.80M & 59.3 & $\pm0.0$ {\scriptsize[$0.0$, $+0.1$]} & 68.7 & 0.132 & 3.4 \\
\textbf{shipped (int8)} & 0.80M & \textbf{59.2} & (release) & 68.5 & 0.137 & 1.5 \\
\bottomrule
\end{tabular}%
}
\caption{Ablations on ParseBench-2, each a swap inside the release
pipeline; $\Delta$ = paired difference to the release with its 95\% bootstrap CI over documents. \emph{OCR models}: other OCR
models in place of ours (PP-OCR through RapidOCR); the last row drops LayoutNet. \emph{LayoutNet variants}: the tables-only model
has no layout heads, so layout falls back to LiteParse's heuristic blocks. \emph{Latin recognizer}: three sizes trained from
scratch on the same data and schedule; the shipped model has the slim architecture.}
\label{tab:ablations}
\end{table*}

\begin{figure}[t]
\centering
\includegraphics[width=\columnwidth]{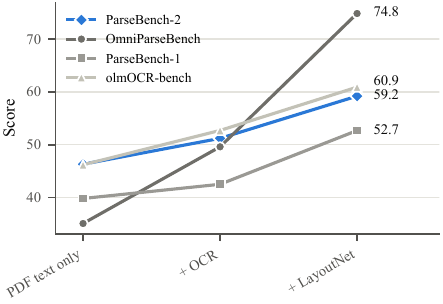}
\caption{Parsing accuracy as OCR and LayoutNet are added to LiteParse's native PDF text extraction.
Each stage includes the preceding components.}
\label{fig:ladder}
\end{figure}

In this section, we perform ablations on several components to determine which parts of
\ocria{} are capacity limited vs. limited by labels. We grow or shrink each model with
the general recipe and keep data the same. We find that \ocria{} is capacity-bound in its Latin recognizer,
while the detector and LayoutNet are not. We also find that each component in \ocria{}
adds step-changes in accuracy, as described in \Cref{fig:ladder}.

\paragraph{Each stage has impact.} \Cref{fig:ladder} adds the components one after the other.
On ParseBench-2 the PDF text layer alone scores 46.3. LayoutNet without OCR raises
that to 53.7, OCR without LayoutNet to 51.2, and both together combine to 59.2. Replacing our OCR models with
Tesseract costs 2.4 points, and dropping OCR entirely costs 5.5 (\cref{tab:ablations}). OCR models
2--9$\times$ larger (PP-OCRv6 small and medium) recover at most 0.9 points.

\paragraph{Each LayoutNet input and head adds accuracy.} \Cref{tab:ablations} swaps LayoutNet
variants into the release pipeline:

\begin{itemize}
  \item \textbf{Layout heads}: replacing LiteParse's heuristic-based blocks with LayoutNet's raises layout 16.7 points from 53.0 to 69.7, while tables drop by 1.8 points.
  \item \textbf{Image branch}: integrating image features adds 1.8 table and 3.0 layout points (+1.1 overall).
\end{itemize}

The image-branch gain is measured from the wider joint model (d256), which on its own is within noise of the narrower one
(58.2 against 58.3).

\paragraph{Filtering heuristic tables.} Enabling the heuristic-table filter in \cref{sec:layoutnet} adds 1.2 points on
ParseBench-2 [+0.9, +1.4], mostly from layout (+6.0). Text, formatting and forms each gain 0.4 points, while tables
lose 0.1. The filter also adds 0.4 points on OmniParseBench, with olmOCR-bench unchanged.

\begin{figure}[t]
\centering
\includegraphics[width=\columnwidth]{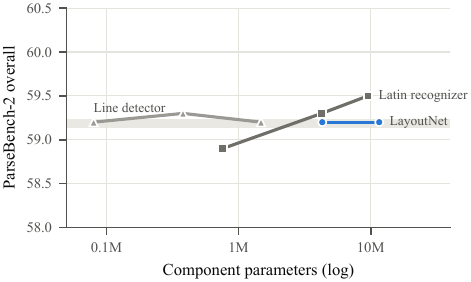}
\caption{ParseBench-2 overall against component size: the line detector, the Latin recognizer,
and LayoutNet, each swept inside the release pipeline. Band: the release
with LayoutNet retrained from three seeds; differences inside it are noise. Only the Latin recognizer rises clearly with size.}
\label{fig:capacity}
\end{figure}

\paragraph{The Latin recognizer is capacity-bound.} The Latin recognizer improves steadily with
size (\cref{tab:ablations,fig:capacity}), at the cost of latency. Trained from scratch on the same data, moving from 0.76M to 9.6M parameters
lowers HierText CER from 0.162 to 0.116 and adds 2.4 ParseBench-2 text points (0.7 overall). The shipped slim model reaches
0.137 at nearly the same size (0.80M, covering 791 characters).
Since the accuracy gains are small overall, we ship the slim model and spend the savings on throughput.

\paragraph{The detector is not capacity-bound.} Swapped into the release package, detectors of 0.08M (shipped),
0.38M and 1.48M parameters all score 59.2--59.3 on ParseBench-2, with text between 68.3 and 68.5. The
shipped detector is 18$\times$ smaller than the largest at no measurable cost.

\paragraph{LayoutNet is also not capacity-bound.} Growing it to 11.6M parameters changes the overall score by $\pm0.0$
[$-0.3$, $+0.2$]: 59.2 against the shipped 4.3M model's 59.2 (\cref{tab:ablations}).
LayoutNet improves with better inputs (the image branch) and better labels (grounded boxes),
but not with size (\cref{tab:ablations}). Future work
would see us improve our labelling pipeline to increase performance at this size, or even look at
smaller models.

\section{Related work}
\label{sec:related}

\paragraph{End-to-end VLM parsers.}
Vision-language models (VLMs) have become a popular approach to end-to-end document parsing, especially for markdown
output. Nougat \citep{nougat} (a Swin Transformer image encoder with an mBART text decoder),
GOT-OCR 2.0 \citep{gotocr2} (a vision encoder and Qwen-0.5B decoder) and olmOCR \citep{olmocr} (a Qwen2-VL-7B fine-tune) were early examples. Recent works
have similar designs while they differ mainly in the specifics around vision encoders, sizes, and training recipes.
OCRia-1 \citep{ocria1}, the teacher for \ocria{}, is a 4B fine-tune of Qwen3.5-4B. Other peers
include Chandra 2 \citep{chandra}, LightOnOCR-3 \citep{lightonocr3}, Infinity-Parser2-Pro \citep{infinityparser},
dots.ocr \citep{dotsocr} and Nemotron 3.5 Parse \citep{nemotronparse}, all being various flavors of text decoders with vision encoders.
VLMs are at the frontier of accuracy, with OCRia-1 scoring 87.0 on ParseBench-2, well above any CPU system (\cref{tab:main_results}).
However, VLM parsers all require a GPU for useful throughput and latency. For example, OCRia-1 parses between 2 and 10 pages per second on a single
B200, depending on request concurrency.

\paragraph{Small CPU OCR pipelines.}
Small detector--recognizer systems sit at the opposite end of the document parsing spectrum. Their output is typically text lines with boxes
rather than markdown, and they run on CPUs at high throughput and low latency. Tesseract
\citep{smith2007tesseract} is among the most popular legacy systems for OCR. Since version 4 it has used an LSTM line recognizer, and it is the default OCR
engine in LiteParse.
One of the most widely used is the PP-OCR series from Paddle \citep{ppocr,paddleocr}. PP-OCRv6 \citep{ppocrv6} pairs DBNet detectors with CTC recognizers, in various
sizes (1.5M to 34.6M parameters), and produces only text and boxes. We use PP-OCRv6 as an
agreement check when building labels (\cref{sec:data}). Swapped into LiteParse in place of our OCR models, its larger
recognizers provide a small increase in accuracy at a large cost in throughput compared to \ocria{} (\cref{tab:efficiency}).

Indra \citep{indra} is the closest comparison to \ocria{}. It has the same overall pipeline shape, containing a text detector,
per-script CTC recognizers behind a script router, and smaller models for page orientation, reading order and math. In
our runs Indra roughly matches \ocria{}'s throughput but trails it on accuracy (\cref{tab:main_results}).
Marker \citep{marker} wraps Indra in an end-to-end converter with additional heuristics, similar to LiteParse with \ocria{}.
Unlike both, \ocria{} is built around LiteParse's spatial projection of the PDF text layer, which it uses as input features. It also
predicts table structure and layout blocks jointly with a single learned model, LayoutNet.

\paragraph{Text-layer converters.}
Text-layer converters such as PyMuPDF4LLM \citep{pymupdf4llm} and pdfplumber \citep{pdfplumber} apply heuristics to the
text embedded in born-digital PDFs. They are the fastest option for those documents, but they have weak or no OCR for
scanned pages. LiteParse \citep{liteparse} also belongs to this category depending on the settings used. It extracts text with PDFium and projects it
onto a grid that preserves the page's spacing. \Cref{fig:ladder} measures what each learned stage of \ocria{}
adds on top of the text layer alone.

\paragraph{Layout analysis and table structure.}
Once text extraction works, the next problem is labelling layout and table structure, which let downstream systems organize and process the
text. DocLayNet \citep{doclaynet} is a core dataset for training layout models. These range from the LayoutLM family
\citep{layoutlm}, which models text and layout together, to purely visual box detectors such as the Heron models
\citep{heron}. Table structure recognition has its own datasets, such as PubTables-1M \citep{pubtables} and FinTabNet
\citep{fintabnet}, and its own models, including TableFormer \citep{tableformer} and the lightweight POTATR
\citep{potatr}. Visual models like Heron and TableFormer detect regions from page pixels. LayoutNet instead labels the
text pieces that LiteParse already has, from either the PDF text layer or OCR. In one pass it assigns them to table cells
and to 13 block classes, and a small image branch adds visual context. Replacing LiteParse's heuristic blocks with
LayoutNet substantially improves layout accuracy (\cref{sec:eval:ablations}).

\paragraph{Benchmarks.}
End-to-end parsing benchmarks are the main way to compare document parsers today. We evaluate on ParseBench-1
\citep{parsebench1}, ParseBench-2 \citep{parsebench2}, OmniParseBench \citep{omniparsebench}, olmOCR-bench
\citep{olmocr,olmocr2} and the BHL IMPACT historical print set \citep{bhl_impact,bhl_leaderboard}
(\cref{sec:eval:protocol}). Their published leaderboards mostly cover VLMs and cloud APIs, and small CPU systems are
often missing from them.
\section{Conclusion and future work}
\label{sec:conclusion}

We presented \ocria{}, an 8.1M-parameter model pipeline for CPU document parsing. It is a pipeline of models made up of a text-line detector, a script router,
per-script recognizers (3.9M parameters in total), and LayoutNet (a 4.3M-parameter joint table and layout model). Training data
is distilled from OCRia-1, using multi-reader agreement to filter and collect training data. With LiteParse as the orchestrator, \ocria{} can replace the
Tesseract default as an opt-in OCR backend, and raises ParseBench-2 from 48.7 to 59.2, OmniParseBench from 42.3 to 74.8, ParseBench-1 from 40.5 to
52.7, olmOCR-bench from 45.7 to 60.9, and lowers BHL's character error rate from 0.275 to 0.048. It does this on CPU at
27.8 pages per second on 16 vCPUs or 426\,ms per page on a single core (\cref{tab:main_results,tab:efficiency}).

Among other parsers that are meant to run on CPU, \ocria{} leads Indra and Marker on ParseBench-2 (by 4.1 and 6.7 points),
ParseBench-1, OmniParseBench and BHL, and ties them on olmOCR-bench. Most of the lead is attributed to tables and layout,
where LayoutNet helps LiteParse reach 59.5 and 72.4 on ParseBench-2. It runs at comparable speed to other CPU pipelines, slightly below Indra's
throughput and slightly above its per-page latency. We showed that each component in the pipeline provides accuracy through ablations, where on
ParseBench-2 the text layer alone scores 46.3 through existing heuristics, while the entire pipeline stacks on up to 13 more points (\cref{fig:ladder}).
The final pipeline reaches about two-thirds of the teacher's ParseBench-2 score (59.2 against 87.0) with roughly 1/500th of its parameters.

\paragraph{Limitations and future work.}
\begin{itemize}
  \item \textbf{Recognition is the main remaining gap.} The Latin recognizer is the one component that improves with size
  (\cref{tab:ablations}), and text recognition is where \ocria{} could gain against competing models.
  Ablations show that a 9.6M model adds 2.4 text points but costs 7$\times$ more CPU per line. Training longer on more
  data moved the shipped slim model further than the larger architectures did, so we expect data (higher quality and more quantity) to drive
  future gains. Handwriting and scripts with little training data remain the weakest aspects as observed in benchmarks.
  \item \textbf{Charts.} \ocria{} has no chart model to process charts into tables, which leaves the Charts dimension near zero
  (3.1) on ParseBench-2, which is the largest single gap to the VLM parsers.
  \item \textbf{Layout grounding and scanned math.} Layout grounding (72.4 on ParseBench-2) still trails the VLM parsers
  (80--89), and equations on scanned pages are not recovered at all (olmOCR-bench old-scans math). Both point to
  LayoutNet's region labels and a math recognizer for image inputs as possible next steps.
\end{itemize}

\bibliographystyle{plainnat}
\bibliography{references}

\appendix{}
\section{Data details}
\label{app:data}

\paragraph{Licensing.} We train on every source except those under non-commercial or no-derivatives licences
(e.g.\ IAM \citep{marti2002iam}, FUNSD \citep{jaume2019funsd}). Pages from the OCRia-1 corpus are dropped when their
licence field matches either restriction or when the corpus marks them excluded.

\paragraph{Benchmark decontamination.} Before any dataset is built, documents are checked against a blocklist of every
benchmark we report, where file hashes and 8-word text shingles over ParseBench-1, ParseBench-2,
OmniParseBench, olmOCR-bench, OpenDataLoader-Bench, and BHL are used for filtering.

\paragraph{Label agreement.} A line from the PDF text layer or from the teacher is kept only when a second reader agrees
with it to within a character error rate of 0.15 (\cref{sec:data}). We found this to be an acceptable deviation while
still capturing enough data to train a capable model.

\paragraph{Synthetic text.} Synthetic pages render Wikipedia sentences in 47 languages (CC BY-SA) with open-licensed fonts,
including handwriting fonts, under random layouts, sizes and ink. Each line's label is its exact source text, and its box
is the rendered ink. Scripts with little real data (Ethiopic, Armenian, Lao, most Indic scripts) rely mostly on these pages
to train the associated recognizers.

\section{Training details}
\label{app:training}

\paragraph{Recognizers.} All recognizers are CTC models over 32-pixel-high line crops: a convolutional stem followed by
token mixers and a linear head over the script's character set. The Latin recognizer (0.80M parameters, 791 classes)
is quantized to int8 (static QDQ with ONNX Runtime). The world (0.98M) and right-to-left recognizers share the slim body.
The CJK recognizer uses a narrower head and a stride of 8 pixels per CTC step, since CJK glyphs are roughly square.
Training uses batch 128 with a cosine learning-rate schedule. Every crop is augmented randomly using crop jitter, perspective and
elastic warps, ruled paper and bleed-through, low resolution, blur, JPEG, noise, faded or heavy ink, binarisation and slight skew.

\paragraph{Detector and router.} The line detector (0.08M parameters) predicts text-line probability maps on pages
resized to a long side of 1600 pixels. Boxes are cut at column gutters and printed column rules. The router
(39k parameters) classifies each line crop into one of four script groups and predicts whether the line is upside down.

\paragraph{LayoutNet.} A 6-layer pre-norm transformer at width 256 over text units, with a strided CNN image branch whose
patch tokens the top layers cross-attend to (4.3M parameters in total). Node, pair and class heads are trained
jointly on table and layout annotations grounded to the units (\cref{sec:data}).

\section{Evaluation details}
\label{app:eval}

\paragraph{Benchmark versions and settings.} ParseBench-2 is the public release v2.07 (\texttt{llamaindex/ParseBench-2}) scored with the official
\texttt{parse-bench} 2.2.0. Its overall score is the mean of the six dimensions.
ParseBench-1 uses the official ParseBench scorer v2.2.0 with the same LiteParse block mapping, and OmniParseBench the official v1.7 harness.
We set \texttt{--keep-headers-footers} for LiteParse on ParseBench-1, ParseBench-2 and OmniParseBench, whose ground truth keeps running
heads, and \texttt{--no-links} everywhere. olmOCR-bench runs with LiteParse's defaults, which strip running heads,
plus \texttt{--no-links}.

\paragraph{ParseBench-2 scoring.} ParseBench-2 uses severity-aware scoring, where a page's score is capped by its
most severe error, meaning a single changed number or broken table relation can cost most of the page.

\paragraph{BHL scoring.} We use the leaderboard's scorer (v3.0) on all 2,165 pages. We report micro-averaged
character error rate on body text using its reading-lane normalization, which ignores case and normalizes ligatures
and the historical long s \citep{bhl_leaderboard}.

\paragraph{Result sources.} All CPU results in \cref{tab:main_results,tab:pb2} are from our runs, using
LiteParse's release build where applicable. For GPU and cloud systems, ParseBench-2 scores are re-scored from
published outputs in ParseBench-2-results with the same official scorer. Other reference scores come from the
OmniParseBench v1.7 leaderboard, the official ParseBench-1 leaderboard, the overall olmOCR-bench results in the
OCRia-1 report, and the BHL leaderboard.

\paragraph{Baseline configurations.} Tesseract 5 is an LSTM line recognizer without a published parameter count
(its English model is about 16\,MB). PP-OCRv6 comes in three sizes, tiny (1.5M parameters), small (7.7M) and medium
(34.6M). Indra is the most similar pipeline to \ocria{}, consisting of a similar set of script recognizers behind
a router, along with satellite models for page orientation, reading order and math, amounting to $\sim$12M parameters
in total. Marker is a pipeline around Indra similar to LiteParse, adding slightly more heuristic processing for documents.

\paragraph{Reproducing the ParseBench-2 rows.} With \texttt{parse-bench} installed and the ocria-nano models pulled
(\texttt{lit models pull ocria-nano}), \texttt{parse-bench run liteparse\_ocria\_nano --dataset parsebench-2} reproduces the
release row; \texttt{indra} and \texttt{marker\_fast} run Indra 1.1.2 and Marker 2.2.0 locally (their CLIs on \texttt{PATH}),
and \texttt{liteparse\_tesseract} the default configuration.

\paragraph{Competitor runs.} Indra 1.1.2 and Marker 2.2.0 (fast mode) run through their own providers on ParseBench-2.
For other evaluation harnesses, we use small wrappers that give them LiteParse's command line.
Indra runs with math enabled, as LiteParse's math
is always on, and with the same per-benchmark header and footer setting as LiteParse. Marker runs with its defaults,
except that its bundled Indra predictor is limited to one ONNX Runtime thread per worker, like every other
configuration in the speed runs.
Indra's OmniParseBench leaderboard entry is 71.4; \cref{tab:main_results} reports our run with the official harness.

\paragraph{Speed protocol.} Throughput is measured on the same 300 ParseBench-1 pages as \cref{tab:efficiency} with 16 single-core
workers in a 16-vCPU container. Latency is one worker on the same pages in four
interleaved batches. Every configuration runs in the same container on each of five hosts (\cref{tab:throughput}).
\Cref{tab:efficiency} reports throughput as mean $\pm$ standard deviation over the five hosts and warm, single-core
latency averaged over hosts.

\begin{table}[t]
\centering\small
\adjustbox{max width=\linewidth}{%
\begin{tabular}{@{}lrrrr@{}}
\toprule
Configuration & ParseBench-2 & OmniParseBench & ParseBench-1 & olmOCR-bench \\
\midrule
PDF text only & 46.3 & 35.1 & 39.81 & 46.2 \\
+ LayoutNet (no OCR) & 53.7 & 58.6 & 49.27 & 52.9 \\
+ OCR (no LayoutNet) & 51.2 & 49.6 & 42.50 & 52.7 \\
\midrule
Release (OCR + LayoutNet) & \textbf{59.2} & \textbf{74.8} & \textbf{52.68} & \textbf{60.9} \\
\bottomrule
\end{tabular}%
}
\caption{Each pipeline stage on every benchmark (the numbers behind \cref{fig:ladder}).}
\label{tab:ladder_full}
\end{table}

\begin{table}[t]
\centering\small
\adjustbox{max width=\linewidth}{%
\begin{tabular}{@{}lrcr@{}}
\toprule
System & CER $\downarrow$ & 95\% CI & Recall \\
\midrule
dots.mocr (VLM) & 0.024 & -- & -- \\
chandra-ocr-2 (VLM) & 0.029 & -- & -- \\
PP-OCRv6-medium & 0.034 & -- & 0.98 \\
\textbf{\ocria{}} & 0.048 & [0.031, 0.062] & 0.93 \\
Indra 1.1.2 & 0.062 & [0.026, 0.098] & 0.94 \\
tesseract-5 & 0.064 & [0.042, 0.079] & 0.92 \\
Marker (fast) & 0.067 & [0.029, 0.108] & 0.94 \\
\bottomrule
\end{tabular}%
}
\caption{BHL IMPACT ground truth, 2,165 historical pages: reading-lane character error rate with 95\% confidence
intervals where we ran the system ourselves.}
\label{tab:bhl_full}
\end{table}

\begin{table}[t]
\centering\small
\adjustbox{max width=\linewidth}{%
\begin{tabular}{@{}llrrrrr@{}}
\toprule
Host & CPU & Tess. & Indra & v6 tiny & v6 small & \ocria{} \\
\midrule
Azure 1 & AMD EPYC, 3.1 GHz & 13.5 & 26.6 & 12.6 & 5.4 & 25.3 \\
Azure 2 & AMD EPYC, 3.0 GHz & 12.6 & 26.0 & 13.0 & 5.4 & 24.7 \\
Azure 3 & AMD EPYC, 3.1 GHz & 13.4 & 27.7 & 13.8 & 5.7 & 25.6 \\
AWS & Intel Xeon, 3.8 GHz & 16.6 & 38.8 & 19.4 & 6.2 & 35.3 \\
GCP & AMD EPYC, 2.4 GHz & 15.0 & 29.0 & 14.6 & 6.0 & 28.3 \\
\midrule
Mean & & 14.2 & 29.6 & 14.7 & 5.7 & 27.8 \\
sd & & 1.6 & 5.3 & 2.7 & 0.4 & 4.4 \\
\bottomrule
\end{tabular}%
}
\caption{Pages per second on 300 random ParseBench-1 pages, 16 single-core workers in a 16-vCPU container. All configurations
run in the same container on each host. Tess.\ = LiteParse default; v6 tiny / small = LiteParse + LayoutNet + PP-OCRv6
through RapidOCR.}
\label{tab:throughput}
\end{table}

\section{Runtime}
\label{app:runtime}

The models run in-process in LiteParse's Rust engine on ONNX Runtime, loaded at run time.
The engine uses only the CPU execution provider, with one intra-op thread pool shared by every model session. The
models download with \texttt{lit models pull}, which pins each file's hash. Installed with ONNX Runtime, the package
takes 59.9\,MB on Linux and 75.3\,MB on macOS.

\onecolumn
\section{Qualitative examples}
\label{app:examples}

\begin{figure}[!htbp]
\centering
\includegraphics[width=\textwidth]{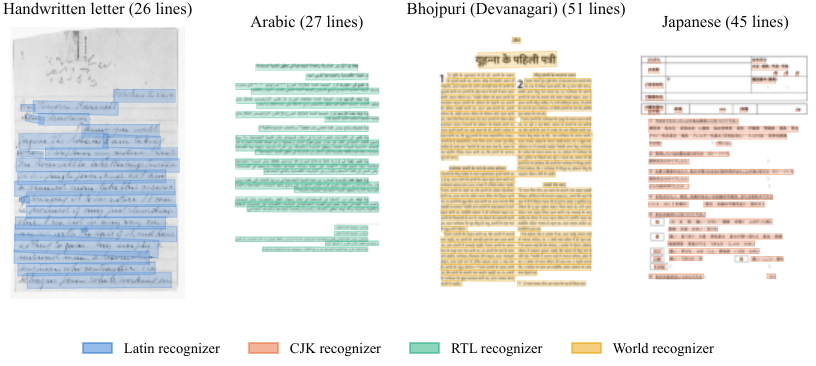}
\caption{Detected lines coloured by the recognizer the script router picks.}
\label{fig:ocr_routing}
\end{figure}

\begin{figure}[!htbp]
\centering
\includegraphics[width=0.85\textwidth]{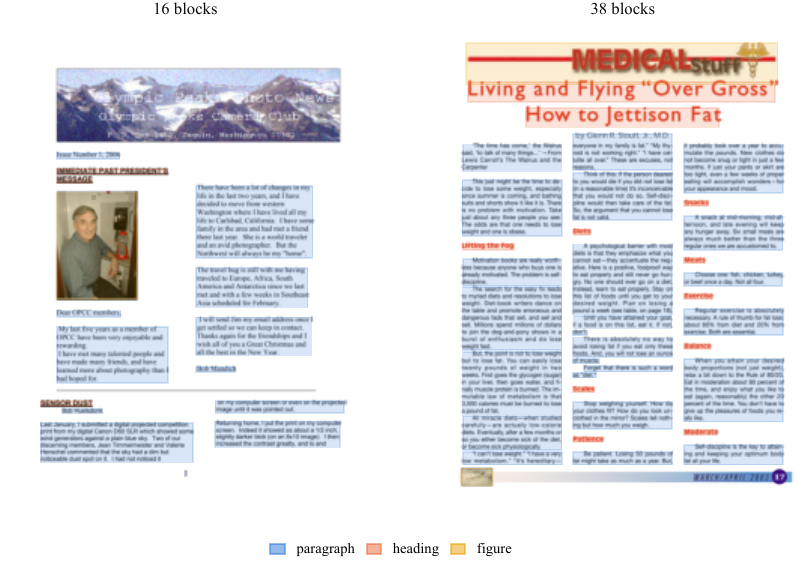}
\caption{LayoutNet layout blocks on two multi-column pages (text layer only).}
\label{fig:layout_blocks}
\end{figure}

\Cref{fig:ocr_routing,fig:layout_blocks} illustrate script routing and layout reconstruction.

\end{document}